\documentclass[letterpaper, 10 pt, conference]{ieeeconf}  

\IEEEoverridecommandlockouts                              

\usepackage{hyperref} 
\usepackage{graphicx}
\usepackage{multirow}
\usepackage{siunitx}

\title{\LARGE \bf

SculptDiff: Learning Robotic Clay Sculpting from Humans with Goal Conditioned Diffusion Policy

}

\author{Alison Bartsch, Arvind Car, Charlotte Avra, and Amir Barati Farimani
\thanks{$^1$A. Bartsch, A. Car, C. Avra and A. B. Farimani are with the Department of Mechanical Engineering at Carnegie Mellon University, Pittsburgh, PA, 15213, USA (e-mail: {\tt\small abartsch@andrew.cmu.edu}, {\tt\small afariman@andrew.cmu.edu})}
\thanks{$^2$A. Bartsch is supported by the Philip and Marsha Dowd Engineering Seed Fund research fellowship}
}

\begin{document}

\maketitle
\thispagestyle{empty}
\pagestyle{empty}

\begin{abstract}
Manipulating deformable objects remains a challenge within robotics due to the difficulties of state estimation, long-horizon planning, and predicting how the object will deform given an interaction. These challenges are the most pronounced with 3D deformable objects. We propose SculptDiff, a goal-conditioned diffusion-based imitation learning framework that works with point cloud state observations to directly learn clay sculpting policies for a variety of target shapes. To the best of our knowledge this is the first real-world method that successfully learns manipulation policies for 3D deformable objects. For sculpting videos and access to our dataset and hardware CAD models, see the project website: \href{https://sites.google.com/andrew.cmu.edu/imitation-sculpting/home}{https://sites.google.com/andrew.cmu.edu/imitation-sculpting/home}

\end{abstract}

\section{INTRODUCTION}
\label{introduction}

Advancements in robotic deformable object manipulation have large-scale implications ranging from manufacturing \cite{kimble2022performance, lv2022dynamic} to surgery \cite{liu2021real}. However, deformable object manipulation remains an open challenge within the robotics field due to the complexities of the interaction between the object and robot, as the object permanently changes shape with each grasp. In this work, we aim to explore the challenges of deformable object manipulation with the task of autonomously sculpting clay. The clay sculpting task is a useful benchmark to investigate methods for deformable objects due to the difficulty of the task itself. Firstly, the deformation behavior is difficult to predict as clay has no underlying structure and can be deformed in all three dimensions. When sculpting clay, the system needs to have a representation of the 3D shape, which poses observation and state representation challenges. Additionally, the system needs to have a sense of the goal shape and execute a sequence of actions that result in this final goal shape. This is particularly challenging as multiple sequences of actions can result in the same final 3D shape, but the ordering of the actions themselves are important, presenting a difficult planning problem.

Within the realm of deformable object manipulation there has been recent success learning and planning with a dynamics model to predict the complicated interactions between a rigid end-effector and a deformable object. However, for more complicated sculpting tasks with 3D objects, planning with a dynamics model can be very time consuming at test time due to the large state and action space \cite{bartsch2023, shi2022, shi2023}. To address this long planning time, we can instead train a policy to go directly from observations to actions. However, due to the high complexity of deformable objects, it is very challenging to train a robust policy for 3D deformable object sculpting in simulation or in the real-world \cite{liu2023softgpt, qi2022}. This motivates our proposed work, where instead we train a policy directly from human demonstrations to avoid the exploration challenges. In this work, we present SculptDiff, a point cloud-based diffusion policy for the clay sculpting task that can successfully sculpt a 3D target shape from only 10 real-world demonstrations. The key contributions of this work are:

\begin{figure}
  \centering
  \includegraphics[width=\linewidth]{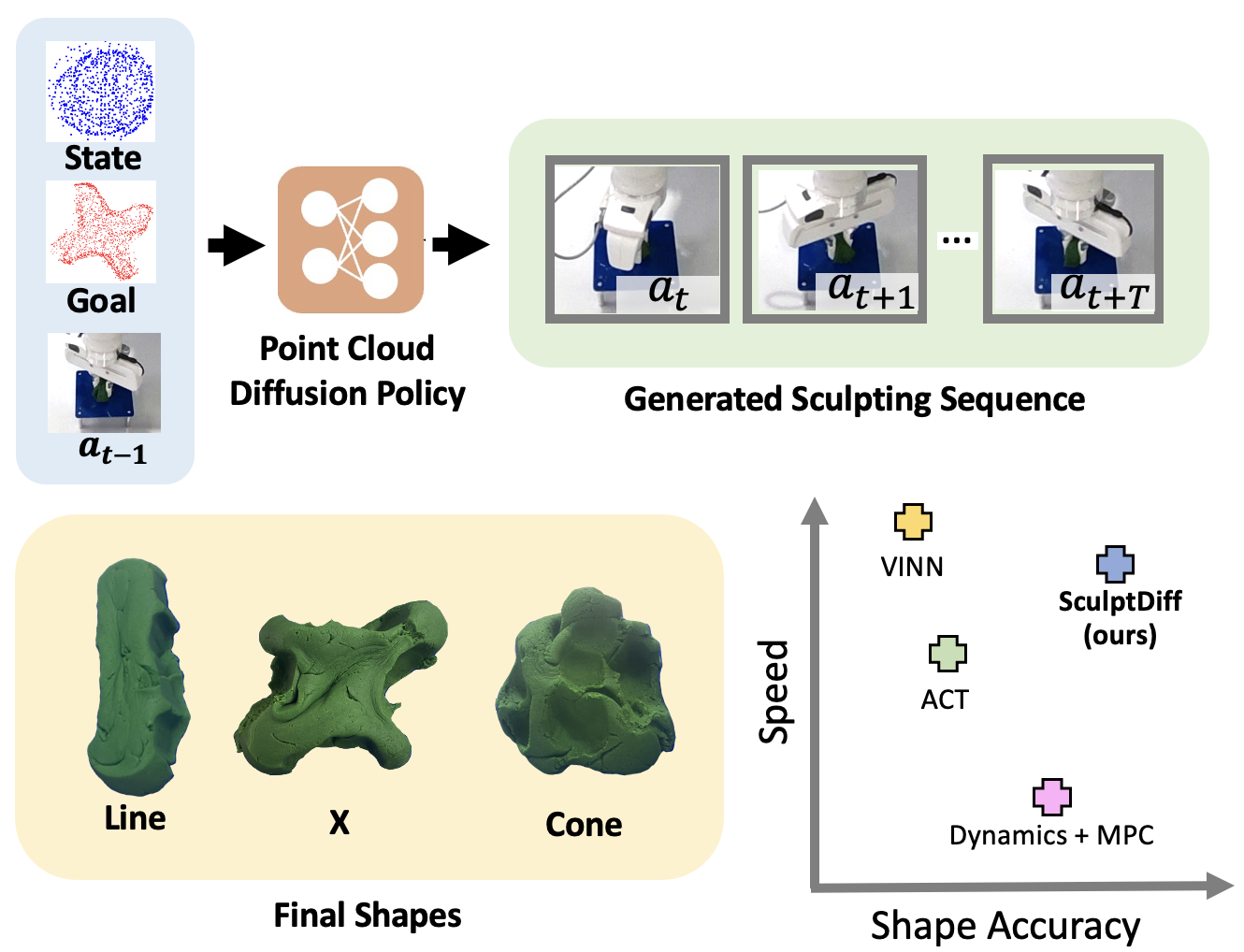}
  \caption{\label{fig:overview} We present a goal-conditioned imitation learning framework for sculpting clay that uses point-cloud state representation. We find that our system is much faster at test time compared to traditional methods planning with a dynamics model. However, our system is more limited in scope as imitation learning does not generalize well to unseen goals.}
  \label{figurelabel}
\end{figure}



\begin{itemize}
    \item To the best of our knowledge we present the first successful imitation learning pipeline to autonomously sculpt a variety of shapes with 3D deformable objects in the real world. 
    \item We propose key modifications of a state-of-the-art imitation learning framework to incorporate 3D data into state representations for 3D deformable object manipulation tasks.
    \item We provide access to a rich dataset of human demonstrations sculpting 3D deformable objects into a variety of shapes in the real world. Additionally, we provide all necessary CAD models to directly recreate our entire experimental setup to improve replicability of results.
    
\end{itemize}

\section{RELATED WORK}

\textbf{Imitation Learning:} Imitation learning involves training a policy from expert demonstrations. In its simplest form of behavior cloning, it is a supervised-learning framework in which a model is trained to directly replicate an expert's behaviors \cite{pomerleau1988alvinn}. There is a wide variety of successful modern behavior cloning frameworks. \cite{pari2021} presents a simple method in which visual observations are clustered and smooth actions are generated using locally weighted regression. With advancements in model architectures, there have been many transformer-based behavior cloning frameworks that have had success on a wide variety of robotic manipulation benchmark tasks \cite{brohan2022, jang2022, shafiullah2022}. While successful, a key challenge of behavior cloning is that mistakes made at previous timesteps compound and lead the robot to experience states out of the training distribution. Building off of the success of transformer frameworks, the action chunking transformer (ACT) \cite{zhao2023} is a transformer-based framework which addresses this challenge of compounding errors by predicting a sequence of actions and averaging over these predictions at each execution step to smooth the overall motion. ACT has since been successfully modified to learn from a single human demonstration \cite{george2023b}, and to incorporate text-based prompts \cite{bharadhwaj2023}. However, due to the temporal averaging of actions, the ACT framework struggles learning a successful manipulation policy when the demonstration trajectories have multi-modal action distributions. Diffusion policy \cite{chi2023} is an imitation learning system that is able to predict a sequence of actions and handle multi-modal action distributions by representing the policy as a conditional denoising diffusion process. All of these imitation learning frameworks are designed primarily for rigid object manipulation with image observations. In this work we present a strategy for behavior cloning with goal conditioning and point cloud observations for deformable object tasks.

\begin{figure*}
  \centering
  \includegraphics[width=0.99\linewidth]{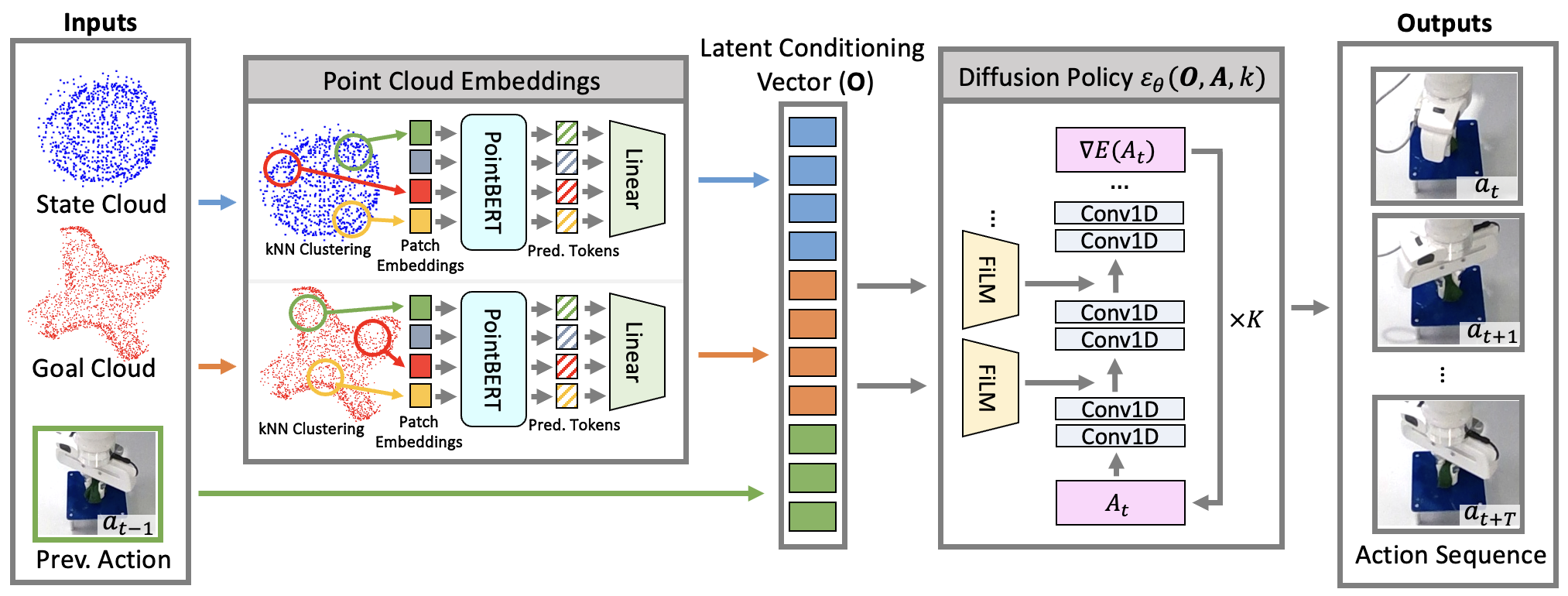}
  \caption{\label{fig:main} The pipeline of SculptDiff. The state and goal point clouds are encoded with PointBERT \cite{yu2022point} and a linear projection head to create a latent conditioning observation along with the previous action executed by the robot. The latent state and goal observations as well as the previous action are the conditioning information used to condition the denoising diffusion process for diffusion policy \cite{chi2023} to generate the predicted action sequence.}
  \label{figurelabel}
\end{figure*}

\textbf{Deformable Objects:} The successful methodologies for handling and shaping deformable objects primarily depend on the properties of the objects themselves, particularly if they are 1D such as cables and rope, 2D such as cloth, or 3D such as clay and dough. While all deformable objects share the key challenges of state estimation and dynamics modeling, the best solutions vary within these different regimes. Within the 1D domain, researchers have found the most success tracking, predicting and modeling behavior leveraging keypoints along the object \cite{huo2022, xiang2023}, or with edge detection \cite{caporali2022}. Works creating simple target shapes out of wire often successfully employ planning strategies without requiring any learning component \cite{zhu2019, yu2023}. However, when the object lacks rigidity, such as rope, a learning component is necessary to successfully create target shapes \cite{zhang2021}. This pattern continues into the 2D domain. As state observation becomes more difficult, rgb inputs may no longer be sufficient. Works addressing the complexities of state observation have trained an image embedding end-to-end with a latent dynamics model \cite{yan2021}, modeled the cloth as a mesh to reason about self-occlusion \cite{huang2022}, used optical flow as an input to reason about cloth motion and structure \cite{weng2022}, or leveraged point cloud observations \cite{lin2022a}. State estimation and dynamics prediction becomes significantly more challenging for 3D deformable objects, making real-world systems particularly difficult. To address this issue, researchers created PlasticineLab \cite{huang2021}, a differentiable physics simulation environment for dough manipulation tasks. PlasticineLab has enabled works to train a dynamics model for dough rolling and squeezing tasks \cite{chen2022}, learn simple abstracted dough manipulation skills from unlabeled human demonstrations \cite{lin2022b, li2023}, and successfully transfer a dough flattening policy from simulation to the real world \cite{qi2022}. However, most real-world systems for 3D deformable object manipulation involve somewhat structured objects, such as toys, sponges, or bottles. While these objects do pose some similar challenges to that of clay or dough, the underlying structure often simplifies the state estimation and dynamics predictions. For example \cite{thach2022} and \cite{shen2023} are able to use partial-view point clouds because the underlying structure of the object can be assumed. One of the few fully real-world approaches for unstructured 3D deformable objects found success with dough cutting and rolling by planning sequences in a latent space \cite{lin2022c}. In this work, we are focusing on the difficulties of manipulating fully unstructured clay in the real world to explore questions of state representation and policy learning challenges for the very complicated task of sculpting.




\textbf{Sculpting Task:} Within the realm of robotic sculpting, there are subtractive strategies in which clay is slowly removed from the starting block to create a target sculpture, and deformation strategies in which the clay is squeezed and deformed to create the final target sculpture. Among the subtractive sculpting works, there have been CNC-style subtractive methods for creating wood sculptures \cite{xuejuan2007}, task planning methods \cite{ma2020}, and human-robot collaboration methods in which a human artist makes small adjustments to the robot path to improve the final shape \cite{zhang2023}. For the purpose of studying the complexities of the interaction between a robot and deformable objects, in this work we are primarily concerned with deformation-based sculpting strategies. Due to the difficulty of this task, there are only a handful of existing works. In \cite{li2022}, researchers use contact-point discovery in to create a variety of shapes out of clay, however all experiments are conducted in simulation. In \cite{shi2022}, researchers present a learned GNN dynamics model and gradient-based planner to sculpt clay into a variety of letters using point cloud inputs. In the follow-up work, this framework is expanded to include a set of dough manipulation tools and a learned tool classifier network \cite{shi2023}. Alternatively, \cite{bartsch2023} leverages pre-trained point cloud embeddings to learn a latent dynamics model for a variety of shape sculpting tasks. To the best of our knowledge, these three works are the only systems that successfully create deformation-based sculptures in the real world. While these results are impressive, they all involve planning with a learned dynamics model which can be quite slow at test time. In this work, we aim to successfully sculpt a set of shapes directly with a learned policy to significantly improve the sculpting speed.


\section{METHOD}

In this work we present SculptDiff, a goal-conditioned imitation learning framework that uses point clouds as state observations for 3D deformation-based clay sculpting. An overview of the pipeline is shown in Figure \ref{fig:main}. Our work consists of an analysis of point clouds as a state representation, the point cloud embedding strategy and latent diffusion policy, and an explanation of the deformation sculpting task.

\subsection{Clay Sculpting Task}

We define the clay sculpting task as applying a sequence of parallel grasp deformations to a piece of clay fixed to the workspace with the goal of replicating a target point cloud. We collect point cloud observations before and after each grasp. The action space is defined as the x, y, z position of the end-effector, the rotation about the z-axis and the distance between the fingertips at the end of the squeeze action. It is important to note that in our current framework there is no notion of forces, however this could be an area to explore in follow-up work. A sculpting trajectory is defined as the sequence of point cloud states and actions given a target shape point cloud, i.e. $T = \{s_0, a_0, ... ,s_N, a_N, s_{N+1} \mid g\}$, where $s$ is the state point cloud, $a$ is the 5D parameterized action, $N$ is the number of grasp actions and $g$ is the goal point cloud. We limit point cloud observations of the clay state to before and after each complete grasp to avoid difficulties in partial point cloud observation when the gripper is occluding the clay. This allows us to relax some assumptions made in other successful sculpting works that require 3D models of the end-effector tools to approximate point cloud completion for occluded observations \cite{shi2022, shi2023}. However, this decision makes the sculpting task more difficult because there is no observational information during the gripper's interaction with the clay. Additionally, this means the state observations are more disparate over time, which can make imitation learning more difficult.



\subsection{Point Cloud State Representation} \label{pcl}

\begin{figure}
  \centering
  \includegraphics[width=0.9\linewidth]{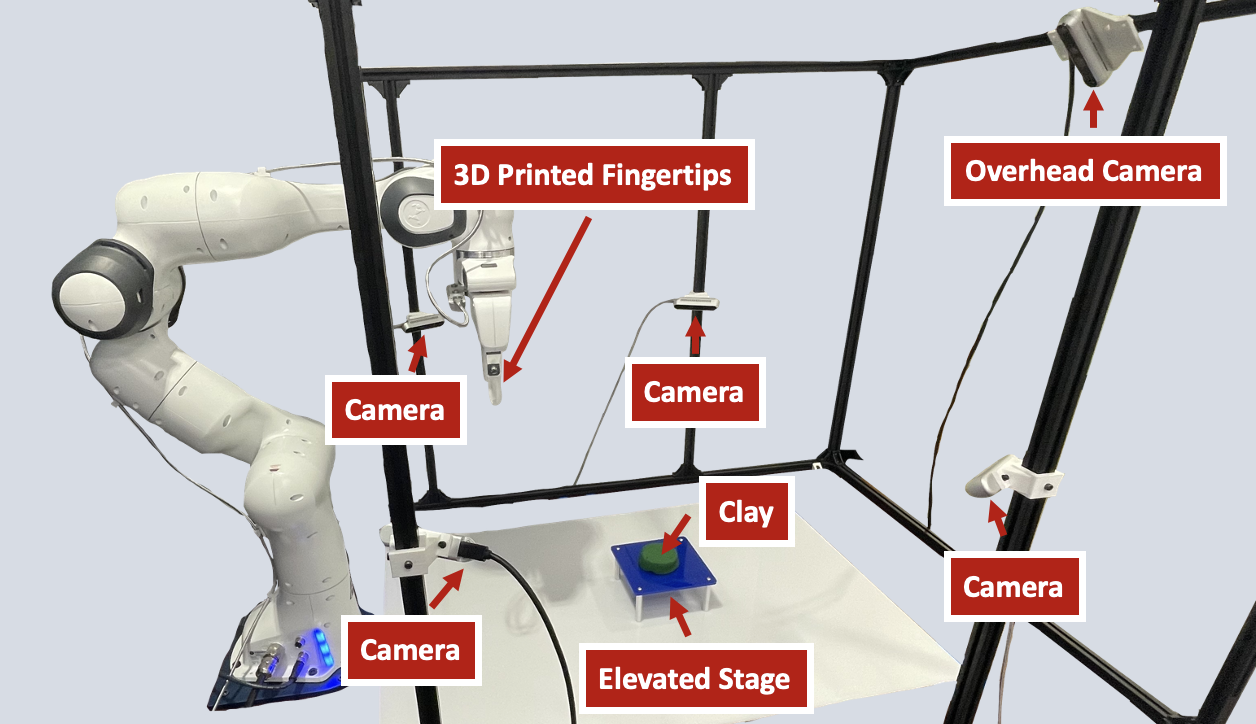}
  \caption{\label{fig:experiment} The experimental setup includes 4 Intel RealSense D415 RGB-D cameras mounted to a camera cage to reconstruct the clay point cloud. An additional Intel RealSense D455 overhead camera is used to record experimental videos. We fit the robot with 3D printed fingertips and an elevated stage similar to those in \cite{bartsch2023}. We assume the clay always remains centered and fixed to the elevated stage throughout the experiments.}
  \label{figurelabel}
\end{figure}

The task of sculpting clay into a target shape requires reasoning about 3D geometry. Thus, we hypothesize that our system requires an observation space that explicitly represents this 3D information. Beyond the fact that we are training a policy for an inherently 3D task, past studies have shown that for general cases using point clouds as observations compared with RGB and RGB-D observations show improvement in robot performance on a variety of manipulation tasks \cite{zhu2024}. Particularly, in \cite{zhu2024} researchers found that beyond increasing success rates, point cloud observations make the learned policies more resilient to environmental variations such as camera pose changes, lighting levels and background appearances. Furthermore, in past work we have demonstrated that pre-trained point cloud embedding models can provide sufficient latent representations of clay point cloud data \cite{bartsch2023}. Beyond these benefits, point clouds as the state observations allows us to augment our demonstration dataset, minimizing the number of hardware demos that are necessary to train a quality sculpting policy. Our augmentation strategy is based on the key assumption that the clay always remains fixed to the elevated stage (shown in Figure \ref{fig:experiment}). For each demonstration trajectory, which is sequence of point cloud observations and grasp actions as well as a target point cloud, we apply a rotational transform about the z-axis in $\ang{1}$ increments to both the state and goal clouds as well as to the parameterized action. This allows us to transform a single human demonstration into 360 varying demonstrations. This augmentation strategy is not easily feasible with image-based observations. Classical image-based augmentation strategies typically involve applying transformations to the images in the form of random shifts, cropping, masking, etc. \cite{yarats2020image}. However these augmentation strategies are designed to make the image embeddings more robust to variations, but do nothing to diversify the action demonstrations themselves. Given that demonstration collection is incredibly time consuming, particularly for a difficult task such as sculpting, reducing the number of real-world demonstrations needed to learn a quality sculpting policy is significant. 

To acquire the full 3D point cloud of the clay state, we use 4 Intel RealSense D415 cameras that are mounted to a camera cage for simple multi-camera calibration. Our physical camera setup is shown in Figure \ref{fig:experiment}. We use the same point cloud processing pipeline as in our previous work \cite{bartsch2023}, in which the raw point clouds from each camera are transformed into the robot coordinate frame and combined together. This combined point cloud of the entire workspace is then cropped based on the known position of the elevated stage. Next, all points are removed if they fall below a hard-coded color thresholding to remove regions of the cloud which are not the clay. In this case we opted to use green clay, but this operation could be used for clay of any color. Finally, we add in a base plan to the clay shell based on the assumption that the clay remains fixed to the table, and uniformly downsample this final cloud to a fixed 2048 points.

\begin{table*}[]
\caption{Overall accuracy of imitation learning frameworks on 'X' shape with point cloud and image state inputs.}\label{tab:general_performance} 
\centering
\begin{tabular}{@{\extracolsep{\fill}}llllllll}
        \hline
        \textbf{Observation} & \textbf{Model} & \textbf{CD $\downarrow$} & \textbf{EMD $\downarrow$} & \textbf{Hausdorff $\downarrow$} & \textbf{\# Grasps} & \textbf{Inference Time (s) $\downarrow$} \\ 
        \hline
        \hline
        \multirow{5}{*}{\textbf{Point Cloud}} & Diffusion & \textbf{0.0073 $\pm$ 0.0007} & \textbf{0.0071 $\pm$ 0.0007} & \textbf{0.0287 $\pm$ 0.0036} & \textbf{6.1 $\pm$ 0.9} & 0.22 $\pm$ 0.03 \\
          & ACT & 0.0077 $\pm$ 0.0007 & 0.0077 $\pm$ 0.0007 & 0.0533 $\pm$ 0.0025 & 8.0 $\pm$ 0.0 & 1.89 $\pm$ 0.75 \\
           & VINN & 0.0156 $\pm$ 0.0017 & 0.0137 $\pm$ 0.0010 & 0.0573 $\pm$ 0.0035 &  9.0 $\pm$ 0 & 0.055 $\pm$ 0.00 \\
            & Heuristic & 0.0087 $\pm$ 0.0004 & 0.0079 $\pm$ 0.0003 & 0.0418 $\pm$ 0.0127 & 10.0 $\pm$ 0.0 & \textbf{0.001 $\pm$ 0.00} \\
            & SculptBot \cite{bartsch2023} & 0.0081 $\pm$ 0.0006 & 0.0082 $\pm$ 0.0008 & 0.0370 $\pm$ 0.0033 &  4.0 $\pm$ 1.10 & 31.27 $\pm$ 2.30 \\
        \hline
        \multirow{3}{*}{\textbf{Image}} & Diffusion & 0.0088 $\pm$ 0.0024 & 0.0119 $\pm$ 0.0052 & 0.0399 $\pm$ 0.0102 & \textbf{5.2 $\pm$ 0.84} & 0.29 $\pm$ 0.01 \\
          & ACT & \textbf{0.0084 $\pm$ 0.0001} & \textbf{0.0092 $\pm$ 0.0003} & \textbf{0.0391 $\pm$ 0.0009} & 10.0 $\pm$ 0.0 & 1.78 $\pm$ 0.29 \\
           & VINN & 0.0140 $\pm$ 0.0005 & 0.0126 $\pm$ 0.0006 & 0.0552 $\pm$ 0.0069 & 10.0 $\pm$ 0 & \textbf{0.0562 $\pm$ 0.0014} \\
        \hline 
        \multirow{2}{*}{\textbf{N/A}} & R. Demo & 0.0075 $\pm$ 0.0004 & 0.0095 $\pm$ 0.0010 & 0.0312 $\pm$ 0.0030 & \textbf{8.0 $\pm$ 0.0} & N/A \\
          & H. Demo & \textbf{0.0059 $\pm$ 0.0006} & \textbf{0.0053 $\pm$ 0.0009} & \textbf{0.0230 $\pm$ 0.0029} & 13.0 $\pm$ 0.9 & N/A \\
        \hline 
    \end{tabular}
\end{table*}

\begin{table}[]
\caption{Performance of SculptDiff trained on single (S) or multiple (M) shape goals with point cloud inputs.}\label{tab:detailed_performance} 
\centering
\begin{tabular}{@{\extracolsep{\fill}}llllll}
        \hline
        \textbf{} & \textbf{Model} & \textbf{CD $\downarrow$} & \textbf{EMD $\downarrow$} & \textbf{\# Grasps} \\ 
        \hline
        \hline
        \multirow{7}{*}{\textbf{X}} & Diff. (S) & \textbf{0.0073 $\pm$ 0.001} & \textbf{0.0071 $\pm$ 0.001} & 6.1 $\pm$ 0.9  \\
          & Diff. (M) & 0.0077 $\pm$ 0.001 & 0.0129 $\pm$ 0.002 & \textbf{5.6 $\pm$ 0.6} \\
           & ACT (S) & 0.0077 $\pm$ 0.001 & 0.0077 $\pm$ 0.001 & 8.0 $\pm$ 0.0 \\
            & VINN (S) & 0.0156 $\pm$ 0.002 & 0.0137 $\pm$ 0.001 & 9.0 $\pm$ 0.0 \\
            & Heuristic & 0.0087 $\pm$ 0.000 & 0.0079 $\pm$ 0.000 & 10.0 $\pm$ 0.0 \\
            & R. Demo & 0.0075 $\pm$ 0.000 & 0.0095 $\pm$ 0.001 & \textbf{8.0 $\pm$ 0.0} \\
            & H. Demo & \textbf{0.0059 $\pm$ 0.001} & \textbf{0.0053 $\pm$ 0.001} & 13.0 $\pm$ 0.9 \\
        \hline
        \multirow{7}{*}{\textbf{Line}} & Diff. (S) & \textbf{0.0045 $\pm$ 0.000} & \textbf{0.0041 $\pm$ 0.000} & 8.0 $\pm$ 0.0 \\
            & Diff. (M) & 0.0064 $\pm$ 0.001 & 0.0085 $\pm$ 0.002 & 6.0 $\pm$ 0.7 \\
          
           & ACT (S) & 0.0159 $\pm$ 0.003 & 0.0169 $\pm$ 0.003 & 8.0 $\pm$ 0.0 \\
            & VINN (S) & 0.0109	 $\pm$ 0.001 & 0.0104 $\pm$ 0.001 & 8.0 $\pm$ 0.0 \\
            & Heuristic & 0.0071 $\pm$ 0.001 & 0.0079 $\pm$ 0.001 & \textbf{2.8 $\pm$ 0.4} \\
            & R. Demo & \textbf{0.0064 $\pm$ 0.004} & 0.0074 $\pm$ 0.004 & \textbf{5.0 $\pm$ 0.0} \\
            & H. Demo & 0.0065 $\pm$ 0.001 & \textbf{0.0065 $\pm$ 0.001} & 9.8 $\pm$ 0.7 \\
        \hline
        \multirow{7}{*}{\textbf{Cone}} & Diff. (S) & 0.0060 $\pm$ 0.001 & \textbf{0.0054 $\pm$ 0.001} & 12.0 $\pm$ 0.0 \\
          & Diff. (M) & \textbf{0.0059 $\pm$ 0.001} & 0.0078 $\pm$ 0.002 & 12.0 $\pm$ 0.0 \\
           & ACT (S) & 0.0070 $\pm$ 0.000 & 0.0079 $\pm$ 0.000 & 12.0 $\pm$ 0.0 \\
            & VINN (S) & 0.0096 $\pm$ 0.001 & 0.0092 $\pm$ 0.001 & 13.0 $\pm$ 0.0 \\
            & Heuristic & 0.0074 $\pm$ 0.000 & 0.0067 $\pm$ 0.000 & \textbf{8.8 $\pm$ 0.8} \\
            & R. Demo & 0.0057 $\pm$ 0.001 & 0.0074 $\pm$ 0.002 & \textbf{10.2 $\pm$ 0.7} \\
            & H. Demo & \textbf{0.0038 $\pm$ 0.001} & \textbf{0.0032 $\pm$ 0.001} & 16.7 $\pm$ 8.5 \\
        \hline 
    \end{tabular}
\end{table}

\subsection{SculptDiff: Point Cloud Diffusion Policy}

In this work, we combine diffusion policy with point cloud state and goal inputs for the robotics sculpting task. An overview of the point cloud diffusion pipeline is shown in Figure \ref{fig:main}. In diffusion policy \cite{chi2023}, the robot policy is represented as a denoising diffusion probabilistic model (DDPM). DDPMs \cite{ho2020denoising} are generative models that iteratively denoise an input sampled from Gaussian noise. One of the key innovations of diffusion policy was incorporating visual observation conditioning in which the DDPM approximates the conditional distribution of $p(\mathbf{A_t} \mid \mathbf{O_t})$, where $\mathbf{A_t}$ is the predicted action sequence and $\mathbf{O_t}$ is the observation the action sequence is conditioned on. In this framework, the action denoising process follows the following equation:

\begin{equation}
    \mathbf{A_t} ^{k-1} = \alpha (\mathbf{A_t} ^k - \gamma \epsilon_{\theta} (\mathbf{O_t}, \mathbf{A_t} ^ k, k) + \mathcal{N} (0, \mathbf{\sigma}^2 \mathbf{I}))
\end{equation}

Where $\mathbf{A_t}^{k}, \mathbf{A_t}^{k-1}, ... , \mathbf{A_t}^{0}$ is the action sequence with decreasing amounts of noise, $k$ is the iteration step,  $\epsilon_{\theta} (\mathbf{O_t}, \mathbf{A_t} ^ k, k)$ is the noise prediction network and $\mathcal{N} (0, \mathbf{\sigma}^2 \mathbf{I})$ is the added Gaussian noise. The parameters $\alpha$, $\gamma$ and $\sigma$ are the noise schedule. To train the model, noise-free action sequences, $\mathbf{A_t}^0$ are sampled from the demonstration dataset, and a random denoising iteration $k$ is selected. The noise prediction network $\epsilon_{\theta}$ is trained to predict the noise in the sampled data with noise added to it. The loss function is shown in the equation below:

\begin{equation}
    Loss = MSE(\epsilon ^k,\epsilon_{\theta} (\mathbf{O_t}, \mathbf{A_t} ^ k + \epsilon^k, k))
\end{equation}

Notice that in this framework, the DDPM is only predicting the action sequence and does not need to predict the next observation state which drastically reduces the dimensions of the vector that we are denoising. This helps speed up the entire denoising process, which makes it more compatible with controlling a robot, as DDPM models can often have slow inference times \cite{li2023diff}. In this work, we are using a 1D CNN-based diffusion policy and modeling the conditional distribution $p(\mathbf{A_t} \mid \mathbf{O_t})$ with feature-wise linear modulation (FiLM) consistent with the recommendations in \cite{chi2023}. 

In the original diffusion policy framework, the observation conditioning $\mathbf{O_t}$ was a vector stacking the flattened latent embeddings of the past $N$ image observations of the scene as well as the robot joint state. In this work, our observation conditioning instead is a learned latent embedding of the clay state and goal point clouds as well as the previous deformation action applied to the clay. To provide a quality latent embedding representing the 3D geometrical information of point clouds, we use PointBERT \cite{yu2022point}, a point transformer encoder, pre-trained on the ShapeNet dataset \cite{chang2015shapenet}. The intuition behind this pre-training step is to ensure our point cloud encoder has learned how to represent a wide variety of 3D shapes before being fine-tuned on a much smaller more limited dataset of human demonstrations. PointBERT is then finetuned end-to-end with the diffusion policy training. PointBERT takes a point cloud, and clusters it into 64 sub-clouds to learn both the overall global geometry as well as the more regional features. The output of PointBERT is $H = \{h_s, h_1, ..., h_g\}$ where $h_s$ represents the global feature and $h_1, ..., h_g$ represents the regional features. While all of this information is very relevant, for the downstream task of learning a policy, we need a much more compact latent representation of the point cloud geometry. Thus, to combine PointBERT with the downstream policy, we add on a two-layer MLP projection head to reduce the latent representation to a compact size of 512. In particular, we combine together the entire global feature $h_s$ and the maxpool of the regional features $h_1, ..., h_g$ before passing this combination through the MLP projection head. See Figure \ref{fig:main} for a visualization of the complete point cloud embedding strategy. We use PointBERT with the MLP projection head to encode the current point cloud observation of the clay as well as the goal point cloud separately into feature vectors of shape 512. Finally, the observation vector $\mathbf{O_t}$ that conditions the diffusion process is the stacked latent representation of the clay state and goal as well as the previous sculpting action applied to the clay. The incorporation of the goal point cloud to condition the policy is a critical component of our methodology due to the rotation augmentations applied to the demonstration dataset. Without conditioning based on the goal, the policy is unable to distinguish between varying rotations and cannot learn a quality sculpting sequence.

\begin{figure*}
  \centering
  \includegraphics[width=\linewidth]{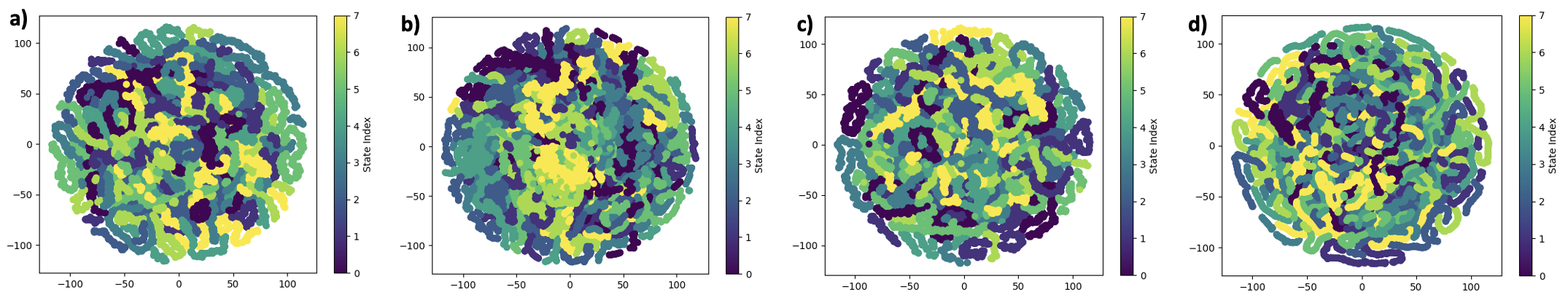}
  \caption{\label{fig:tsne} The TSNE embeddings for PointBERT with different training strategies on the X demonstration dataset. The colorbar of state index indicates the state number ranging from 0 to 7 for the states in each demonstration trajectory for the X shape. a) PointBERT pre-trained on ShapeNet, b) PointBERT finetuned with diffusion policy, c) PointBERT finetuned with ACT policy, and d) PointBERT finetuned with VINN policy.}
  \label{figurelabel}
\end{figure*}

\begin{figure*}
  \centering
  \includegraphics[width=\linewidth]{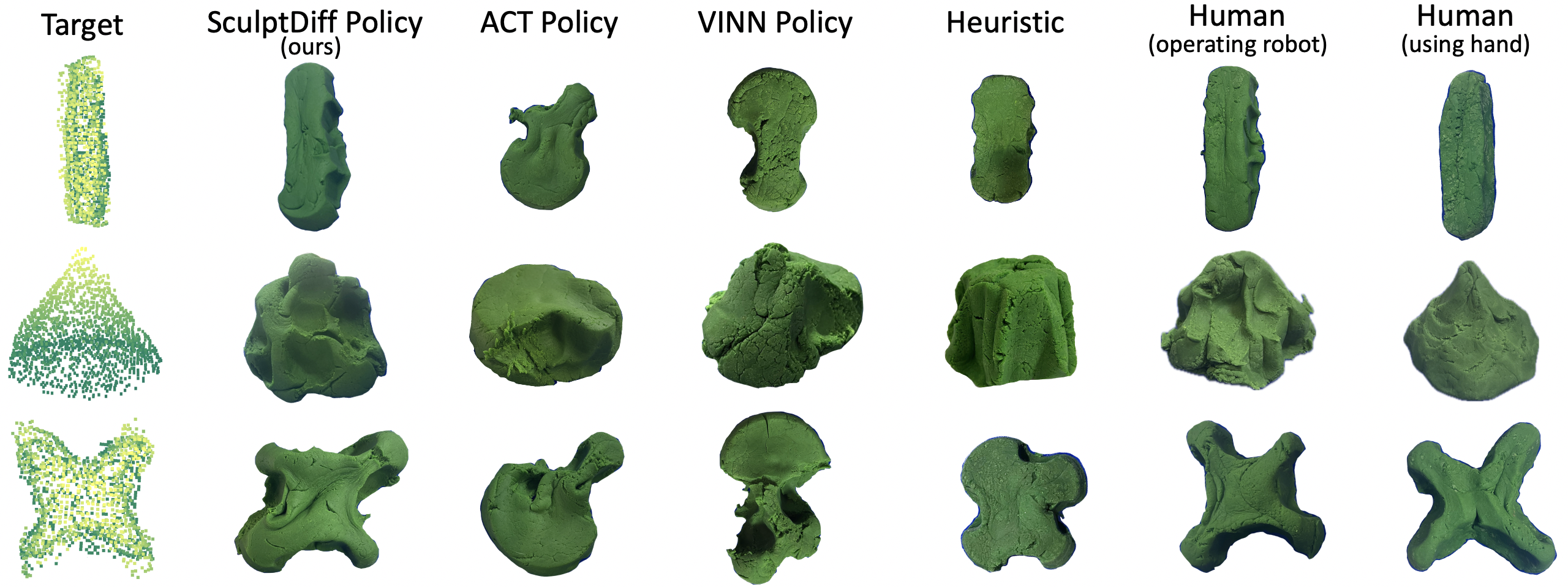}
  \caption{\label{fig:shapes} The final shapes created by the policies trained with point cloud inputs for a single shape goal. For the target point cloud (representing the shapes created by the human oracle using hands) on the left-most column, the lightness of each point is correlated with the point's z-value to visualize depth. While both human oracles create the best shapes, point cloud diffusion policy is able to successfully create the closest matches to the human demonstrations.}
  \label{figurelabel}
\end{figure*}

\section{EXPERIMENTS AND RESULTS}


The human demonstration dataset is collected using kinesthetic teaching in which the expert physically moves the robot to sculpt the clay. We collect 10 demonstration trajectories for each target sculpting shape ('X', 'Line' and 'Cone'), recording the point cloud state of the clay before and after each grasp. We chose these sculpting targets because we believe they allow us to explore a variety of different sculpting behaviors while limiting the amount of time consuming hardware experiments. 'Line' requires the policy to maintain consistent rotation and position about two axes while varying position along the third, 'X' requires a very particular sequence of grasps varying position and rotation along the table plane, while 'Cone' requires reasoning about these same pose variations along with how to vary the gripper height and distance between fingertips. The experimental setup is visualized in Figure \ref{fig:experiment}. We apply the rotation transformation as described in section \ref{pcl}, to our train dataset with an 80/20 split of the raw demonstrations to expand our training dataset to 2880 demonstrations per target shape. To further explore the use of point cloud inputs for imitation learning tasks, we combine the same proposed point cloud embedding strategy with two other state-of-the-art imitation learning frameworks, ACT \cite{zhao2023} and VINN \cite{pari2021}. Similar to that of diffusion policy, these two frameworks were originally designed for image inputs and embed the observations into a latent representation with ResNet encoders. Thus, for both ACT and VINN we are able to directly use the same point cloud embedding and goal conditioning strategy we proposed for diffusion policy. We compare the performance of these models with a series of baselines to evaluate SculptDiff. In the following sections, we separate the discussion of results by the key questions we hope to explore. The numerical results of the sculpting tasks are shown in Tables \ref{tab:general_performance} and \ref{tab:detailed_performance} in which we report the Chamfer Distance (CD), Earth Mover's Distance (EMD) and Hausdorff Distance between the final clay point cloud and the target shape point cloud. We ran each policy 5 times for each shape target and report the mean and standard deviation across experiments. In addition to quantitative similarity, we show the final shapes created by our system compared to a variety of baselines in Figure \ref{fig:shapes}. It is important to note that in this work our target shape point clouds are created by humans using both hands to freely create the simple sculptures. This is a stark difference with past works \cite{bartsch2023, shi2022, shi2023} in which the goal point clouds were generated through a human operating the robot to ensure the targets were possible to create with the robot hardware setup. This makes our shape goals more difficult, and the point cloud distance metrics reflect this. We find the key to the success of SculptDiff is both the access to 3D information via point cloud observations and goals as well as the stochastic nature of diffusion policy itself.

\subsection{How does the point cloud input influence sculpting performance compared to an image input?}

     To fully evaluate the efficacy of our proposed approach for incorporating point clouds as state observations for the sculpting task, we compare the point cloud-based imitation learning models to their original image-based counterparts. In particular, we evaluate the point cloud versus image variants for diffusion policy, ACT and VINN. However, we could not directly use images, as our original dataset consists of only 10 demonstrations which is far too few to train a successful policy. Reiterating one of the strengths of point clouds as compared to images, to create our 'image' dataset we applied the same rotation augmentation strategy to the dense point cloud of the clay on the elevated stage (retaining the RGB information), and converted this cloud back into an image with a fixed camera angle. While this is not a complete representation of what a true image-based sculpting dataset would be, we argue this approximation will still allow us to explore some of the tradeoffs between point cloud and image-based policies. In addition to these image-based baselines we compare SculptDiff to a heuristic baseline as well as two human oracle variants of a person operating the robot and a person using their hand to create the target shapes. The idea of the heuristic baseline is to present a simple framework to solve the sculpting challenge with a learning-free alternative. The heuristic baseline includes Radial and Linear modes, catering to different target shapes. Radial mode centers the gripper above the clay, rotating it as needed, while Linear mode maintains a fixed gripper rotation along the x-axis. For both modes, the current and goal point clouds are divided into equal segments, and the radius of gyration is compared for opposite pairs of these segments to determine the region of the clay with the highest difference between the current and goal point clouds. The gripper moves to the squeeze region and the gripper width is set proportional to the radius of gyration difference.

    From our experiments, the SculptDiff policy outperforms all baselines in terms of CD, EMD, Hausdorff and visual shape quality for all goals. Image-based diffusion policy struggled to create quality 'X' shapes supporting our argument that 3D information is vital for this task. Particularly, the policy struggled to predict accurate z-positions for the end-effector as well as quality sequences of actions dependent on the current observations. Additionally, the inaccuracies in gripper positions at the beginning caused the subsequent clay states to deviate greatly from the original demonstrations, further confusing the image-based policy. Both ACT and VINN struggled with the sculpting task regardless of the state observation. This is likely because both frameworks can struggle with multimodality in demonstrations, and the sculpting task is highly multimodal. There are multiple action sequences that can create a 3D shape, and our demonstrations reflect this. ACT and VINN, both deterministic policies, often get stuck in modes repeating similar grasp actions in perpetuity. This due to the rigidity of these policies, where if the first action is not ideal, and the subsequent states deviate more from the training demonstrations and the models fail to handle the compounding errors, a common issue with imitation learning. For both ACT and VINN, the algorithmic techniques to handle these compounding errors involves averaging, with temporal aggregation for ACT and kernel-averaging the actions of clustered states for VINN. However, this averaging scheme is not compatible with our action parameterization, as averaging two different grasps may result in a third with a substantially different deformation behavior than the original two, thus deviating further from the training states. To ensure we were evaluating these algorithms in the best light for this application, for our experiments we did not use temporal aggregation for ACT and limited the number of nearest neighbors for VINN. While this improved overall performance by reducing the errors caused by the averaging mechanisms, the challenges of compounding errors, particularly with a deterministic policy remained for both. In contrast to ACT and VINN, diffusion policy is stochastic and able to successfully capture the distribution of grasps along the clay over time. Based on the results of our experiments we attribute the success of SculptDiff to both the stochastic representation of actions as well as the point cloud state and goal representations.   

\subsection{How do the policies trained on multiple shape goals and those trained on individual goals compare in behaviors?}

The goal-conditioning component is crucial for training a sculpting policy given our rotation augmentation strategy, as the behavior cloning model needs to have the rotated goal point cloud as input to discern the various rotated demonstration trajectories. However in addition to this, we investigate if goal conditioning allows us to train a single policy on multiple shape goals at once. The performances for these variations are reported for the 'X', 'Line' and 'Cone' shapes in Table \ref{tab:detailed_performance}. A single policy learning multiple varying action sequences for different goals is a stretch task, and multi-goal point cloud diffusion policy is able to maintain successful sculpting performance for the 'Line' and 'Cone' while struggling with the 'X' goal. The multiple goals SculptDiff policy struggles the most with the 'X' likely because maintaining perfect rotation is critical to a good final performance, and confusing the correct rotation even slightly leads to large ongoing inaccuracies in the final shape.

\subsection{How does the point cloud embedding change when finetuning PointBERT end-to-end with different policies?}

We can visualize the point cloud embedding space with t-SNE plots, shown in Figure \ref{fig:tsne}. From this visualization of the embedded latent space for the X demonstration trajectories, we can see that the diffusion policy fine-tuning strategy is the most successful at clustering more similar states closer together. This clustering behavior of distinguishing different states along the demonstration trajectories is correlated with the respective policies' performance on the sculpting tasks, with point cloud diffusion policy the most successful. This is meaningful, as one would expect a successful sculpting policy to correctly learn to identify how close to the goal shape the current state is, and diffusion policy is by far the most successful at the sculpting task.





\subsection{How does our system compare to human performance?}

We compare human performance to SculptDiff in Tables \ref{tab:general_performance} and \ref{tab:detailed_performance}, and visualize the shapes in Figure \ref{fig:shapes}. We conducted experiments with a human creating goal shapes using kinesthetic teaching as well as experiments with a human using their hand to create the sculptures. This allows us to explore the shape quality drop due to a rigid parallel gripper and difficulty controlling the robot. SculptDiff is able to successfully create shape sculptures of 'Line', 'Cone' and 'X' both visually and based on our point cloud similarity metrics. However, both the human operating the robot and the human using their hand outperform SculptDiff in terms of final shape quality and quantitative similarity metrics. The stark differences between the two human oracle experiments highlight key areas for future exploration. In particular, we found that a human using their own hand tends to apply many more smaller changes to the clay than when using the more cumbersome and indelicate robotic setup. Additionally, the CD, EMD, and Hausdorff metrics do not always capture the nuances of overall shape quality. This poses the open question of developing better shape quality metrics for the 3D sculpting task, perhaps exploring more semantic representations of similarity. Furthermore, our current hardware framework makes some shapes more challenging to create than others. The 'Cone' shape in particular was difficult because the current fingertip shape was not as helpful for creating a smooth cone surface as compared to a human hand. For creating more fine-grained sculptures, tools or a softer finger may be necessary. 

\section{CONCLUSION}

In this work, we present SculptDiff, the first imitation learning policy to successfully create a set of 3D clay sculptures entirely in the real world. Through our experiments, we demonstrate the value of leveraging 3D state representations, in this case point clouds, as well as the importance of a stochastic policy for the complex multi-modal task of sculpting. We provide access to the demonstration dataset as well as 3D models to directly recreate the entire hardware setup to encourage replicability. Our evaluation of sculpting quality compared to human baselines has demonstrated a clear need for further exploration of improving hardware to allow for finer changes to be applied to the clay as well as the development of semantic-based 3D shape similarity metrics.














\bibliographystyle{IEEEtran}
\bibliography{references}

\begin{thebibliography}{10}
\providecommand{\url}[1]{#1}
\csname url@samestyle\endcsname
\providecommand{\newblock}{\relax}
\providecommand{\bibinfo}[2]{#2}
\providecommand{\BIBentrySTDinterwordspacing}{\spaceskip=0pt\relax}
\providecommand{\BIBentryALTinterwordstretchfactor}{4}
\providecommand{\BIBentryALTinterwordspacing}{\spaceskip=\fontdimen2\font plus
\BIBentryALTinterwordstretchfactor\fontdimen3\font minus \fontdimen4\font\relax}
\providecommand{\BIBforeignlanguage}[2]{{%
\expandafter\ifx\csname l@#1\endcsname\relax
\typeout{** WARNING: IEEEtran.bst: No hyphenation pattern has been}%
\typeout{** loaded for the language `#1'. Using the pattern for}%
\typeout{** the default language instead.}%
\else
\language=\csname l@#1\endcsname
\fi
#2}}
\providecommand{\BIBdecl}{\relax}
\BIBdecl

\bibitem{kimble2022performance}
K.~Kimble, J.~Albrecht, M.~Zimmerman, and J.~Falco, ``\href{https://www.frontiersin.org/articles/10.3389/frobt.2022.999348/full}{Performance measures to benchmark the grasping, manipulation, and assembly of deformable objects typical to manufacturing applications},'' \emph{Frontiers in Robotics and AI}, vol.~9, p. 999348, 2022.

\bibitem{lv2022dynamic}
N.~Lv, J.~Liu, and Y.~Jia, ``\href{https://ieeexplore.ieee.org/stamp/stamp.jsp?arnumber=9714152&casa_token=hxQf6jtuBF4AAAAA:XI5DxBZZ113xH6rr3LPO6IfaoEdTiH82rs_OdLxgSiQok8tUL8myBfh7i_4CurEtEIP8QJ_b8g}{Dynamic modeling and control of deformable linear objects for single-arm and dual-arm robot manipulations},'' \emph{Transactions on Robotics}, vol.~38, no.~4, pp. 2341--2353, 2022.

\bibitem{liu2021real}
F.~Liu, Z.~Li, Y.~Han, J.~Lu, F.~Richter, and M.~C. Yip, ``\href{https://ieeexplore.ieee.org/stamp/stamp.jsp?arnumber=9561177&casa_token=ZW9xEVKE6coAAAAA:Y77Sp7YZ19pltTO6YPfqj6eP4e7EePvbeqSmSliajJunTNcUIrrZ_nz-I8F_TFBgYN1bp_87fw}{Real-to-sim registration of deformable soft tissue with position-based dynamics for surgical robot autonomy},'' in \emph{International Conference on Robotics and Automation}.\hskip 1em plus 0.5em minus 0.4em\relax IEEE, 2021, pp. 12\,328--12\,334.

\bibitem{bartsch2023}
A.~Bartsch, C.~Avra, and A.~B. Farimani, ``\href{https://arxiv.org/pdf/2309.08728.pdf}{SculptBot: Pre-Trained Models for 3D Deformable Object Manipulation},'' \emph{arXiv preprint arXiv:2309.08728}, 2023.

\bibitem{shi2022}
H.~Shi, H.~Xu, Z.~Huang, Y.~Li, and J.~Wu, ``\href{https://journals.sagepub.com/doi/pdf/10.1177/02783649231219020}{RoboCraft: Learning to see, simulate, and shape elasto-plastic objects in 3D with graph networks},'' \emph{The International Journal of Robotics Research}, p. 02783649231219020, 2023.

\bibitem{shi2023}
H.~Shi, H.~Xu, S.~Clarke, Y.~Li, and J.~Wu, ``\href{https://arxiv.org/pdf/2306.14447.pdf}{RoboCook: Long-Horizon Elasto-Plastic Object Manipulation with Diverse Tools},'' \emph{arXiv preprint arXiv:2306.14447}, 2023.

\bibitem{liu2023softgpt}
J.~Liu, Z.~Li, W.~Lin, S.~Calinon, K.~C. Tan, and F.~Chen, ``\href{https://ieeexplore.ieee.org/stamp/stamp.jsp?arnumber=10341846}{Softgpt: Learn goal-oriented soft object manipulation skills by generative pre-trained heterogeneous graph transformer},'' in \emph{International Conference on Intelligent Robots and Systems}.\hskip 1em plus 0.5em minus 0.4em\relax IEEE, 2023, pp. 4920--4925.

\bibitem{qi2022}
C.~Qi, X.~Lin, and D.~Held, ``\href{https://ieeexplore.ieee.org/stamp/stamp.jsp?arnumber=9830873}{Learning closed-loop dough manipulation using a differentiable reset module},'' \emph{Robotics and Automation Letters}, vol.~7, no.~4, pp. 9857--9864, 2022.

\bibitem{pomerleau1988alvinn}
D.~A. Pomerleau, ``\href{https://proceedings.neurips.cc/paper/1988/file/812b4ba287f5ee0bc9d43bbf5bbe87fb-Paper.pdf}{ALVINN: An autonomous land vehicle in a neural network},'' \emph{Advances in neural information processing systems}, vol.~1, 1988.

\bibitem{pari2021}
J.~Pari, N.~M. Shafiullah, S.~P. Arunachalam, and L.~Pinto, ``\href{https://arxiv.org/pdf/2112.01511.pdf}{The surprising effectiveness of representation learning for visual imitation},'' \emph{arXiv preprint arXiv:2112.01511}, 2021.

\bibitem{brohan2022}
A.~Brohan, N.~Brown, J.~Carbajal, Y.~Chebotar, J.~Dabis, C.~Finn, K.~Gopalakrishnan, K.~Hausman, A.~Herzog, J.~Hsu \emph{et~al.}, ``\href{https://arxiv.org/pdf/2212.06817.pdf}{Rt-1: Robotics transformer for real-world control at scale},'' \emph{arXiv preprint arXiv:2212.06817}, 2022.

\bibitem{jang2022}
E.~Jang, A.~Irpan, M.~Khansari, D.~Kappler, F.~Ebert, C.~Lynch, S.~Levine, and C.~Finn, ``\href{https://proceedings.mlr.press/v164/jang22a/jang22a.pdf}{Bc-z: Zero-shot task generalization with robotic imitation learning},'' in \emph{Conference on Robot Learning}.\hskip 1em plus 0.5em minus 0.4em\relax PMLR, 2022, pp. 991--1002.

\bibitem{shafiullah2022}
N.~M. Shafiullah, Z.~Cui, A.~A. Altanzaya, and L.~Pinto, ``\href{https://proceedings.neurips.cc/paper_files/paper/2022/file/90d17e882adbdda42349db6f50123817-Paper-Conference.pdf}{Behavior Transformers: Cloning $ k $ modes with one stone},'' \emph{Advances in neural information processing systems}, vol.~35, pp. 22\,955--22\,968, 2022.

\bibitem{zhao2023}
T.~Z. Zhao, V.~Kumar, S.~Levine, and C.~Finn, ``\href{https://arxiv.org/pdf/2304.13705.pdf}{Learning fine-grained bimanual manipulation with low-cost hardware},'' \emph{arXiv preprint arXiv:2304.13705}, 2023.

\bibitem{george2023b}
A.~George and A.~B. Farimani, ``\href{https://arxiv.org/pdf/2309.10175.pdf}{One ACT Play: Single Demonstration Behavior Cloning with Action Chunking Transformers},'' \emph{arXiv preprint arXiv:2309.10175}, 2023.

\bibitem{bharadhwaj2023}
H.~Bharadhwaj, J.~Vakil, M.~Sharma, A.~Gupta, S.~Tulsiani, and V.~Kumar, ``\href{https://arxiv.org/pdf/2309.01918.pdf}{Roboagent: Generalization and efficiency in robot manipulation via semantic augmentations and action chunking},'' \emph{arXiv preprint arXiv:2309.01918}, 2023.

\bibitem{chi2023}
C.~Chi, S.~Feng, Y.~Du, Z.~Xu, E.~Cousineau, B.~Burchfiel, and S.~Song, ``\href{https://arxiv.org/pdf/2303.04137.pdf}{Diffusion policy: Visuomotor policy learning via action diffusion},'' \emph{arXiv preprint arXiv:2303.04137}, 2023.

\bibitem{yu2022point}
X.~Yu, L.~Tang, Y.~Rao, T.~Huang, J.~Zhou, and J.~Lu, ``\href{https://arxiv.org/abs/2111.14819}{Point-bert: Pre-training 3d point cloud transformers with masked point modeling},'' in \emph{Conference on Computer Vision and Pattern Recognition}, 2022, pp. 19\,313--19\,322.

\bibitem{huo2022}
S.~Huo, A.~Duan, C.~Li, P.~Zhou, W.~Ma, H.~Wang, and D.~Navarro-Alarcon, ``\href{https://ieeexplore.ieee.org/stamp/stamp.jsp?arnumber=9730102}{Keypoint-based planar bimanual shaping of deformable linear objects under environmental constraints with hierarchical action framework},'' \emph{Robotics and Automation Letters}, vol.~7, no.~2, pp. 5222--5229, 2022.

\bibitem{xiang2023}
J.~Xiang, H.~Dinkel, H.~Zhao, N.~Gao, B.~Coltin, T.~Smith, and T.~Bretl, ``\href{https://ieeexplore.ieee.org/stamp/stamp.jsp?arnumber=10214157}{TrackDLO: Tracking Deformable Linear Objects Under Occlusion with Motion Coherence},'' \emph{Robotics and Automation Letters}, 2023.

\bibitem{caporali2022}
A.~Caporali, K.~Galassi, R.~Zanella, and G.~Palli, ``\href{https://ieeexplore.ieee.org/stamp/stamp.jsp?arnumber=9830852}{FASTDLO: Fast deformable linear objects instance segmentation},'' \emph{Robotics and Automation Letters}, vol.~7, no.~4, pp. 9075--9082, 2022.

\bibitem{zhu2019}
J.~Zhu, B.~Navarro, R.~Passama, P.~Fraisse, A.~Crosnier, and A.~Cherubini, ``\href{https://ieeexplore.ieee.org/stamp/stamp.jsp?arnumber=8851170}{Robotic manipulation planning for shaping deformable linear objects withenvironmental contacts},'' \emph{Robotics and Automation Letters}, vol.~5, no.~1, pp. 16--23, 2019.

\bibitem{yu2023}
M.~Yu, K.~Lv, C.~Wang, M.~Tomizuka, and X.~Li, ``\href{https://ieeexplore.ieee.org/stamp/stamp.jsp?arnumber=10160264}{A coarse-to-fine framework for dual-arm manipulation of deformable linear objects with whole-body obstacle avoidance},'' in \emph{International Conference on Robotics and Automation}.\hskip 1em plus 0.5em minus 0.4em\relax IEEE, 2023, pp. 10\,153--10\,159.

\bibitem{zhang2021}
W.~Zhang, K.~Schmeckpeper, P.~Chaudhari, and K.~Daniilidis, ``\href{https://ieeexplore.ieee.org/stamp/stamp.jsp?arnumber=9560955}{Deformable linear object prediction using locally linear latent dynamics},'' in \emph{International Conference on Robotics and Automation}.\hskip 1em plus 0.5em minus 0.4em\relax IEEE, 2021, pp. 13\,503--13\,509.

\bibitem{yan2021}
W.~Yan, A.~Vangipuram, P.~Abbeel, and L.~Pinto, ``\href{https://proceedings.mlr.press/v155/yan21a/yan21a.pdf}{Learning predictive representations for deformable objects using contrastive estimation},'' in \emph{Conference on Robot Learning}.\hskip 1em plus 0.5em minus 0.4em\relax PMLR, 2021, pp. 564--574.

\bibitem{huang2022}
Z.~Huang, X.~Lin, and D.~Held, ``\href{https://arxiv.org/pdf/2206.02881.pdf}{Mesh-based dynamics with occlusion reasoning for cloth manipulation},'' \emph{arXiv preprint arXiv:2206.02881}, 2022.

\bibitem{weng2022}
T.~Weng, S.~M. Bajracharya, Y.~Wang, K.~Agrawal, and D.~Held, ``\href{https://proceedings.mlr.press/v164/weng22a/weng22a.pdf}{Fabricflownet: Bimanual cloth manipulation with a flow-based policy},'' in \emph{Conference on Robot Learning}.\hskip 1em plus 0.5em minus 0.4em\relax PMLR, 2022, pp. 192--202.

\bibitem{lin2022a}
X.~Lin, Y.~Wang, Z.~Huang, and D.~Held, ``\href{https://proceedings.mlr.press/v164/lin22a/lin22a.pdf}{Learning visible connectivity dynamics for cloth smoothing},'' in \emph{Conference on Robot Learning}.\hskip 1em plus 0.5em minus 0.4em\relax PMLR, 2022, pp. 256--266.

\bibitem{huang2021}
Z.~Huang, Y.~Hu, T.~Du, S.~Zhou, H.~Su, J.~B. Tenenbaum, and C.~Gan, ``\href{https://arxiv.org/pdf/2104.03311.pdf}{Plasticinelab: A soft-body manipulation benchmark with differentiable physics},'' \emph{arXiv preprint arXiv:2104.03311}, 2021.

\bibitem{chen2022}
S.~Chen, Y.~Liu, S.~W. Yao, J.~Li, T.~Fan, and J.~Pan, ``\href{https://ieeexplore.ieee.org/stamp/stamp.jsp?arnumber=9833281}{Diffsrl: Learning dynamical state representation for deformable object manipulation with differentiable simulation},'' \emph{Robotics and Automation Letters}, vol.~7, no.~4, pp. 9533--9540, 2022.

\bibitem{lin2022b}
X.~Lin, Z.~Huang, Y.~Li, J.~B. Tenenbaum, D.~Held, and C.~Gan, ``\href{https://arxiv.org/pdf/2203.17275.pdf}{Diffskill: Skill abstraction from differentiable physics for deformable object manipulations with tools},'' \emph{International Conference on Learning Representations}, 2022.

\bibitem{li2023}
S.~Li, Z.~Huang, T.~Chen, T.~Du, H.~Su, J.~B. Tenenbaum, and C.~Gan, ``\href{https://arxiv.org/pdf/2304.03223.pdf}{DexDeform: Dexterous Deformable Object Manipulation with Human Demonstrations and Differentiable Physics},'' \emph{International Conference on Learning Representations}, 2023.

\bibitem{thach2022}
B.~Thach, B.~Y. Cho, A.~Kuntz, and T.~Hermans, ``\href{https://ieeexplore.ieee.org/stamp/stamp.jsp?arnumber=9812215}{Learning visual shape control of novel 3D deformable objects from partial-view point clouds},'' in \emph{International Conference on Robotics and Automation}.\hskip 1em plus 0.5em minus 0.4em\relax IEEE, 2022, pp. 8274--8281.

\bibitem{shen2023}
B.~Shen, Z.~Jiang, C.~Choy, S.~Savarese, L.~J. Guibas, A.~Anandkumar, and Y.~Zhu, ``\href{https://journals.sagepub.com/doi/pdf/10.1177/02783649231191222}{Action-conditional implicit visual dynamics for deformable object manipulation},'' \emph{The International Journal of Robotics Research}, p. 02783649231191222, 2023.

\bibitem{lin2022c}
X.~Lin, C.~Qi, Y.~Zhang, Z.~Huang, K.~Fragkiadaki, Y.~Li, C.~Gan, and D.~Held, ``\href{https://openreview.net/pdf?id=tyxyBj2w4vw}{Planning with spatial-temporal abstraction from point clouds for deformable object manipulation},'' in \emph{Conference on Robot Learning}, 2022.

\bibitem{xuejuan2007}
N.~Xuejuan, L.~Jingtai, S.~Lei, L.~Zheng, and C.~Xinwei, ``\href{https://ieeexplore.ieee.org/stamp/stamp.jsp?tp=&arnumber=4522463}{Robot 3D sculpturing based on extracted NURBS},'' in \emph{International Conference on Robotics and Biomimetics}.\hskip 1em plus 0.5em minus 0.4em\relax IEEE, 2007, pp. 1936--1941.

\bibitem{ma2020}
Z.~Ma, S.~Duenser, C.~Schumacher, R.~Rust, M.~B{\"a}cher, F.~Gramazio, M.~Kohler, and S.~Coros, ``\href{https://dl.acm.org/doi/pdf/10.1145/3424630.3425415}{Robotsculptor: Artist-directed robotic sculpting of clay},'' in \emph{ACM Symposium on Computational Fabrication}, 2020, pp. 1--12.

\bibitem{zhang2023}
M.~Zhang, Z.~Cheng, S.~T.~R. Shiu, J.~Liang, C.~Fang, Z.~Ma, and S.~J. Wang, ``\href{https://research.polyu.edu.hk/en/publications/cosculpt-an-ai-embedded-human-robot-collaboration-system-for-scul}{CoSculpt: An AI-Embedded Human-Robot Collaboration System for Sculptural Creation},'' in \emph{Human Systems Engineering and Design: Future Trends and Applications}.\hskip 1em plus 0.5em minus 0.4em\relax AHFE International, 2023.

\bibitem{li2022}
S.~Li, Z.~Huang, T.~Du, H.~Su, J.~B. Tenenbaum, and C.~Gan, ``\href{https://arxiv.org/pdf/2205.02835.pdf}{Contact points discovery for soft-body manipulations with differentiable physics},'' \emph{International Conference on Learning Representations}, 2022.

\bibitem{zhu2024}
H.~Zhu, Y.~Wang, D.~Huang, W.~Ye, W.~Ouyang, and T.~He, ``\href{https://arxiv.org/abs/2402.02500}{Point Cloud Matters: Rethinking the Impact of Different Observation Spaces on Robot Learning},'' \emph{arXiv preprint arXiv:2402.02500}, 2024.

\bibitem{yarats2020image}
D.~Yarats, I.~Kostrikov, and R.~Fergus, ``\href{https://openreview.net/pdf?id=GY6-6sTvGaf}{Image augmentation is all you need: Regularizing deep reinforcement learning from pixels},'' in \emph{International conference on learning representations}, 2020.

\bibitem{ho2020denoising}
J.~Ho, A.~Jain, and P.~Abbeel, ``\href{https://proceedings.neurips.cc/paper/2020/file/4c5bcfec8584af0d967f1ab10179ca4b-Paper.pdf}{Denoising diffusion probabilistic models},'' \emph{Advances in neural information processing systems}, vol.~33, pp. 6840--6851, 2020.

\bibitem{li2023diff}
X.~Li, Y.~Liu, L.~Lian, H.~Yang, Z.~Dong, D.~Kang, S.~Zhang, and K.~Keutzer, ``\href{https://openaccess.thecvf.com/content/ICCV2023/papers/Li_Q-Diffusion_Quantizing_Diffusion_Models_ICCV_2023_paper.pdf}{Q-Diffusion: Quantizing Diffusion Models},'' in \emph{International Conference on Computer Vision}, October 2023, pp. 17\,535--17\,545.

\bibitem{chang2015shapenet}
A.~X. Chang, T.~Funkhouser, L.~Guibas, P.~Hanrahan, Q.~Huang, Z.~Li, S.~Savarese, M.~Savva, S.~Song, H.~Su \emph{et~al.}, ``\href{https://arxiv.org/pdf/1512.03012.pdf}{Shapenet: An information-rich 3d model repository},'' \emph{arXiv preprint arXiv:1512.03012}, 2015.

\end{thebibliography}

\end{document}